\pdfoutput=1

\documentclass[11pt]{article}

\usepackage[final]{acl}

\usepackage{times}
\usepackage{latexsym}

\usepackage[T1]{fontenc}

\usepackage[utf8]{inputenc}

\usepackage{microtype}

\usepackage{inconsolata}

\usepackage{graphicx}

\usepackage{underscore}
\usepackage{array}
\newcolumntype{P}[1]{>{\centering\arraybackslash}p{#1}}
\usepackage{amsmath,amssymb,amsfonts}
\usepackage{color}
\usepackage{booktabs}
\usepackage{enumerate}
\usepackage{multicol, multirow}
\usepackage{arydshln}
\usepackage[english]{babel}
\usepackage[autostyle, english=american]{csquotes}
\MakeOuterQuote{"}
\usepackage{subcaption}
\usepackage{soul}
\usepackage{mwe}
\usepackage{wrapfig}
\usepackage{siunitx}
\usepackage{fix-cm}

\title{JudgeRank: Leveraging Large Language Models for Reasoning-Intensive Reranking}

\author{Tong Niu, Shafiq Joty, Ye Liu, Caiming Xiong, Yingbo Zhou, Semih Yavuz \\ 
Salesforce AI Research \\
\texttt{\{tniu,sjoty,yeliu,cxiong,yingbo.zhou,syavuz\}@salesforce.com}
}

\begin{document}
\maketitle

\begin{abstract}
Accurate document retrieval is crucial for the success of retrieval-augmented generation (RAG) applications, including open-domain question answering and code completion. 
While large language models (LLMs) have been employed as dense encoders or listwise rerankers in RAG systems, they often struggle with reasoning-intensive tasks because they lack nuanced analysis when judging document relevance.
To address this limitation, we introduce \textsc{JudgeRank}, a novel agentic reranker that emulates human cognitive processes when assessing document relevance. Our approach consists of three key steps: (1) \textit{query analysis} to identify the core problem, (2) \textit{document analysis} to extract a query-aware summary, and (3) \textit{relevance judgment} to provide a concise assessment of document relevance.
We evaluate \textsc{JudgeRank} on the reasoning-intensive BRIGHT benchmark, demonstrating substantial performance improvements over first-stage retrieval methods and outperforming other popular reranking approaches. 
In addition, \textsc{JudgeRank} performs on par with fine-tuned state-of-the-art rerankers on the popular BEIR benchmark, validating its zero-shot generalization capability.
Through comprehensive ablation studies, we demonstrate that \textsc{JudgeRank}'s performance generalizes well across LLMs of various sizes while ensembling them yields even more accurate reranking than individual models.
\end{abstract}
\section{Introduction}
\label{sec:intro}
Passage reranking is a critical component in modern information retrieval systems, designed to refine results obtained from efficient first-stage retrieval methods such as BM25~\citep{robertson1995okapi,robertson2009probabilistic}. By narrowing down the pool of candidate documents, reranking substantially improves the quality of downstream tasks, such as retrieval-augmented generation or RAG~\citep{LewisRAG}. Two primary approaches have emerged to address the reranking task.
The first category comprises encoding-based approaches~\citep{nogueira2019passage,gao2021rethink}, which encode queries and documents into fixed-size embedding vectors. These methods use either cosine similarity as a score function or directly output a score from the model~\citep{nogueira-etal-2020-document,zhuang2023rankt5}. While highly efficient, these approaches face several limitations. One major challenge is their inflexibility in defining relevance, making it difficult to accommodate diverse retrieval objectives (e.g., finding \textit{supporting} vs. \textit{refuting} evidence). Moreover, encoding-based models heavily rely on manual supervision signals due to the discrepancy between LLM pretraining and reranking objectives, limiting their ability to generalize to new domains or models~\citep{nguyen2016ms,izacard2022unsupervised}. 

Most recently, utilizing Large Language Models (LLMs) for document reranking has led to promising progress in addressing some of these challenges, owing to their superior capabilities in language understanding, generation, interaction, and reasoning~\citep{ouyang2022training}. 
These approaches utilize an LLM either as a pointwise judge~\citep{10.1145/3626772.3657951} or a listwise reranker~\citep{sun-etal-2023-chatgpt,zhuang2024setwise}. While these approaches allow for flexible definition of document relevance and support zero-shot operation, they still require the model to make decisions without intermediate analyses. Consequently, they fall short in scenarios requiring complex reasoning~\citep{su2024bright}, hampering both performance and interpretability. Moreover, listwise rerankers face significant computational challenges due to context length constraints, often compromising on individual document length when processing multiple documents simultaneously.

To bridge this gap, we propose \textsc{JudgeRank}, a novel zero-shot pointwise reranker tailored for reasoning-intensive text retrieval tasks. 
Inspired by Chain-of-Thought~\citep{wei2022chain} and LLM-as-a-Judge~\citep{zheng2023judging} methods, \textsc{JudgeRank} utilizes highly generalizable prompts to guide instruction-tuned LLMs through explicit reasoning steps before arriving at a final judgment.

\begin{figure*}[t]
\centering
\includegraphics[width=0.98\textwidth]{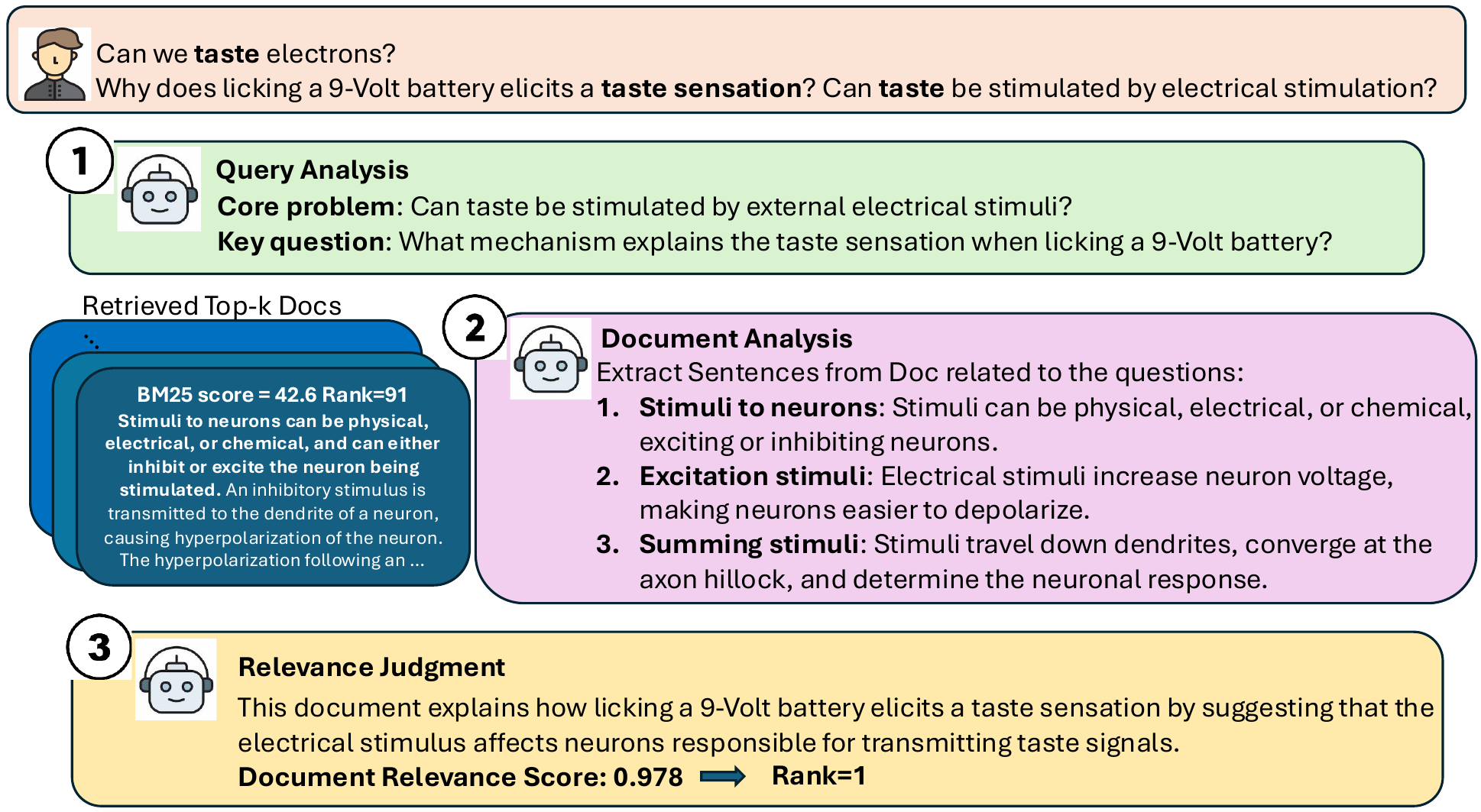}
\caption{A step-by-step illustration of how \textsc{JudgeRank} arrives at the final judgment through query and document analyses. The query analysis identifies the core problem being asked, while the document analysis extracts relevant sentences from the document based on the query. This is a real example from the Biology task in the BRIGHT evaluation benchmark.}
\label{fig:judge-rank}
\end{figure*}

Figure~\ref{fig:judge-rank} illustrates a real example of how our model works on the Biology dataset in the BRIGHT (Benchmark for Reasoning-Intensive Generative Retrieval Tasks) benchmark~\citep{su2024bright}. Specifically, our reranker first prompts the LLM to identify the core problem in the query, allowing it to focus on the central question while filtering out irrelevant context. Next, the model produces an extractive summary for each of the candidate documents and explains how it addresses the query. Finally, the model makes a relevance judgment based on the previous analyses. This process closely mimics how humans approach questions: by first skimming the document, identifying relevant parts, and then carefully reading these parts to obtain an answer~\citep{masson1983conceptual}.
This structured pipeline enables \textsc{JudgeRank} to transcend surface-level lexical matching, leveraging deeper semantic understanding to improve reranking accuracy.

We evaluate \textsc{JudgeRank} on the recently constructed BRIGHT benchmark, widely regarded as one of the most challenging retrieval evaluation datasets. Despite the poor performance of state-of-the-art text embedding models and rerankers on this benchmark, our method achieves significant improvements over all existing baselines and secures the top position on the BRIGHT benchmark leaderboard among $89$ models, surpassing the previous best model by a significant margin ($9$ points).\footnote{\url{https://brightbenchmark.github.io/}}
Additionally, we demonstrate that \textsc{JudgeRank} readily generalizes to other popular retrieval benchmarks such as BEIR and performs competitively with state-of-the-art rerankers. 
We also analyze the complementarity of models at different scales by investigating the alignment of their ranking decisions. 
Contrary to our expectations, we observe that models of different sizes demonstrate a surprisingly orthogonal behavior on their relevance judgments. 
Leveraging this key finding, we propose a simple ensembling strategy that allows us to combine multiple models flexibly and demonstrate that it translates to considerable performance gains on the final ranking.

\section{Method}
\label{sec:method}

\begin{figure*}[t]
    \centering
    \includegraphics[width=0.98\textwidth]{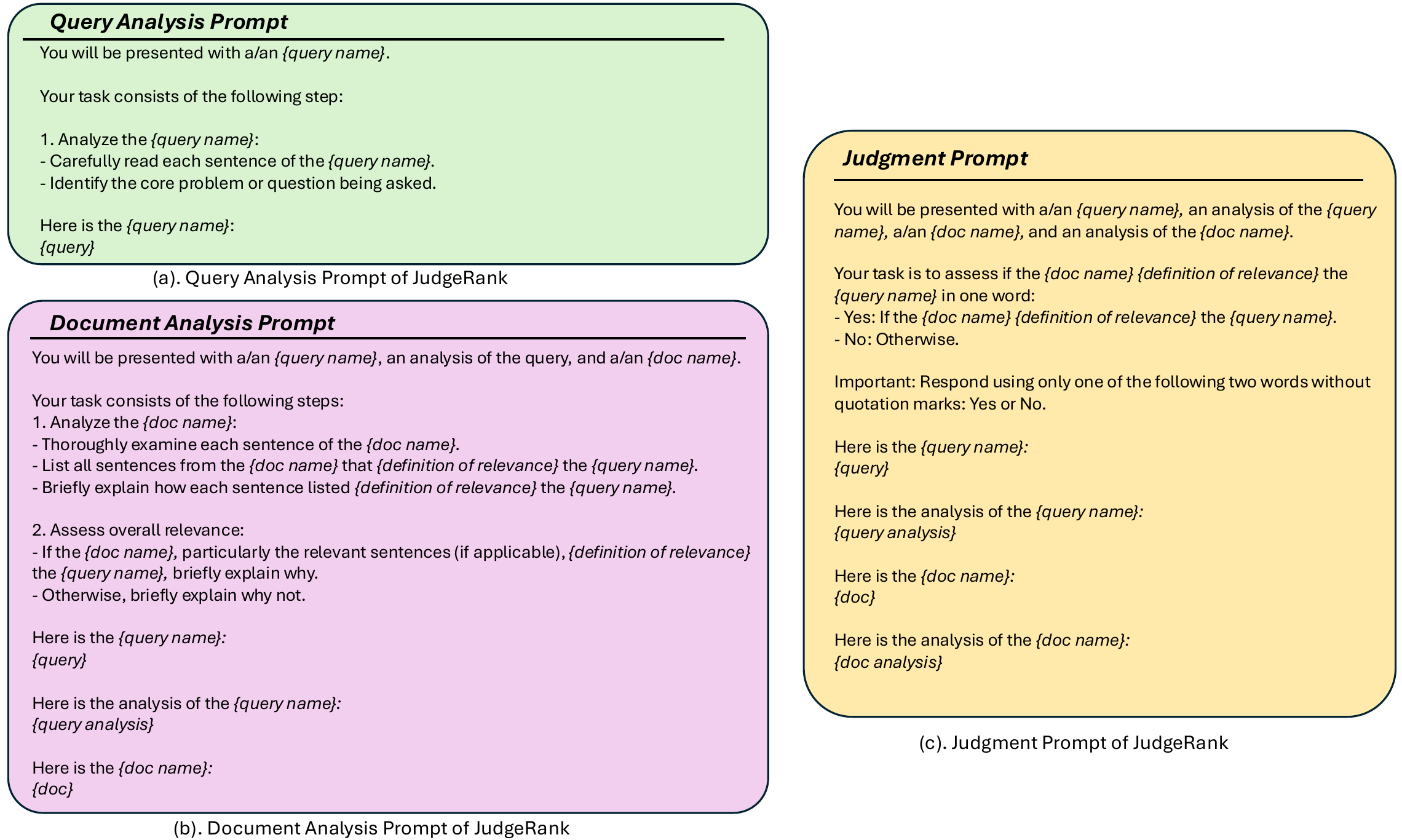}
    \caption{(a) Prompt to analyze query, where \texttt{\{query name\}} (e.g., "Biology post") and \texttt{\{query\}} are placeholders for the query type and content. (b) Prompt for analyzing a document, where \texttt{\{doc name\}} (e.g., "document") and \texttt{\{doc\}} are placeholders for the document type and content. (c) Prompt for making the final one-word relevance judgment.}
    \label{fig:judge prompt}
\end{figure*}

\subsection{Agentic steps}
Mimicing human cognitive process, \textsc{JudgeRank} consists of three main steps: \textit{Query Analysis}, \textit{Document Analysis}, and \textit{Relevance Judgment}. The prompt templates for these steps are illustrated in Figure~\ref{fig:judge prompt}.

\paragraph{Query analysis}
The query analysis prompt (Figure~\ref{fig:judge prompt} (a).) directs the LLM to analyze the query by identifying the core problem being asked. Note that this prompt only depends on the query so that we can generate the query analyses separately and store them. Since the number of queries $n_q$ is usually much smaller than that of documents $n_d$ ($n_q \ll n_d$), we can afford to use a more expensive LLM (e.g., GPT-4) to handle this important step, and leave the other steps to relatively smaller LMs.

\paragraph{Document analysis}
The document analysis prompt (Figure~\ref{fig:judge prompt} (b).) asks the LLM to output an extractive summary of the document that helps answer the query, and assess the overall relevance of the document based on the summary.

\paragraph{Relevance judgment}
The judgment prompt (Figure~\ref{fig:judge prompt} (c).) asks the model to make a one-word judgment, either "Yes" or "No". We isolate this step to make it easier to ensemble with different judgment prompts or models. 

\subsection{Generalizability of the prompts}
All three steps consume natural language as input and generate a response, making it more flexible to transfer and stack across different LLMs. The templates also show that the prompts are highly generalizable: to adapt them to a new reranking task, one only needs to replace the query name, the document name, and the relation between them. Leveraging LLMs in a zero-shot setting allows us to flexibly define the relation between the query and the document. This flexibility is important because the user may define either "document that supports a query" or "document that refutes a query" as the relation, which are opposites of each other. Encoding-based models usually cannot achieve such behavior zero-shot because most of them use cosine similarity or metrics alike to represent "relevance". One way encoding-based models could achieve such flexibility is through extensive fine-tuning. However, this requires additional training data and introduces new model parameters, potentially causing an unintended distribution shift. Similarly, models like \textsc{RankZephyr} also finetunes the LLM to output document ids, thus suffering from a distribution shift as well.

\subsection{Methodology of scoring documents}
\label{subsec:scoring}
\paragraph{Discrete version}
The discrete version creates a binary partition between accepted (when the model outputs a "Yes") and rejected (when the model outputs a "No") documents, maintaining the first-stage retrieval ranking within each category. More specifically, let $D = \{d_1, d_2, \ldots, d_k\}$ be an ordered list of top-$k$ documents ranked by the first-stage retrieval model. Let $D_y$ and $D_n$ be a partition of $D$, such that

\begin{align}
    D_y \cup D_n &= D \\
    D_y \cap D_n &= \emptyset
\end{align}

, where $D_y$ is the set of documents that the reranker judged as relevant and $D_n$ is the set of documents that the reranker judged as non-relevant. Let $R$ be the reranking function which maps each document $d$ to its rank (lower rank means "more relevant"), then

\begin{equation}
    \forall d \in D_y \ \text{and} \ d' \in D_n, \ R(d) < R(d')
\end{equation}

, and for the relative ranking within each partition,

{
    \small
    \begin{equation}
        \forall d_i, d_j \in D_y, \ R(d_i) < R(d_j) \iff R_0(d_i) < R_0(d_j)
    \end{equation}
}

, where $R(d)$ is the final rank of document $d$ and $R_0(d)$ is the rank of $d$ from the first-stage retrieval. The same applies to $D_n$.

While straightforward, this approach is sensitive to prompt wording and relies heavily on first-stage retrieval performance.

\paragraph{Continuous version}
The continuous version addresses the limitations of the discrete version by using the probability of the "Yes" judgment $p_y$ and the probability of the "No" judgment $p_n$ to construct a complete ranking. This is similar to the relevance generation approach proposed in~\citet{liang2023holistic}. The scoring function $S$ is defined by normalizing the probabilities between $p_y$ and $p_n$ as follows:

\begin{equation}
    S(d) = \frac{p_y}{p_y + p_n}
\end{equation}

This normalization step is necessary because the combined probability mass allocated to $p_y$ and $p_n$ is not always a fixed value across different documents. Without normalization, the $p_y$ values for different documents would not be directly comparable.

The final ranking $D_\text{R}$ is then defined as $D_\text{R} = \{d_1, d_2, \ldots, d_k\}$ such that 

{
    \small
    \begin{equation}
        \forall i, j \in \{1, 2, \ldots, k\}, \ i < j \iff S(d_i) > S(d_j)
    \end{equation}
}

This continuous version provides a more fine-grained ranking compared to the binary partition, as it utilizes the full range of probabilities output by the LLM.

\paragraph{Hybrid version}
Additionally, we explore a variant of the continuous version where the final score is computed by taking a weighted sum of the probability score $S_\text{prob}$ and the BM25 score $S_{\text{BM25}}$. More specifically, the final score is computed by

\begin{equation}
    S = \alpha S_\text{prob} + S_{\text{BM25}}
\end{equation}

, where $S_{\text{BM25}}$ is the score provided by BM25 in the first-stage retrieval, and $\alpha$ is the relative weight of the probability score. We set $\alpha = 100$ in this work to bring $S_\text{prob}$ to the same scale as $S_{\text{BM25}}$. This version leverages model ensembling to consider both reasoning and surface-level matching, thus marrying the benefits of both approaches. Unless otherwise specified, we use this setting to compute the final score throughout the paper. In the ablation studies, we compare these three settings and show their relative performances.
\section{Experimental Setup}
\label{sec:experimental-setup}

\subsection{Datasets}
\paragraph{BRIGHT}
We use the BRIGHT benchmark~\citep{su2024bright} to assess the performance of our reranker. BRIGHT is specifically designed to evaluate text retrieval systems on complex, reasoning-intensive queries that go beyond simple keyword matching. The benchmark comprises $1\text{,}398$ real-world queries spanning diverse domains, including economics, psychology, robotics, math, and software engineering. These queries are carefully curated to represent challenging scenarios that require deep understanding and reasoning to identify relevant documents. We use this dataset to evaluate our approach because unlike traditional benchmarks that focus on simple information-seeking tasks, BRIGHT queries require complex reasoning to determine document relevance, making it an excellent tool for evaluating advanced retrieval systems in realistic scenarios. The benchmark has also been validated to be robust against potential data leakage, maintaining its effectiveness even when benchmark documents have been included in model training data.

Because of its challenging nature, state-of-the-art retrieval models have shown significantly lower performance on BRIGHT compared to other benchmarks~\citep{su2024bright}. For example, the leading model on the MTEB leaderboard~\citep{muennighoff2022mteb} achieves an nDCG@10 of only $18.0$ on BRIGHT, compared to $59.0$ on other benchmarks. The GPT-4 listwise reranker also only improves around $2$ points on nDCG@10 on top of the BM25 first-stage retrieval, while Gemini~\citep{team2023gemini} features less improvement than that. The cross-encoder reranker MiniLM~\citep{wang2020minilm} even significantly underperforms the BM25 baseline.

\paragraph{BEIR}
To test the generalizability of our approach, we also evaluate on the BEIR benchmark~\citep{thakur2021beir}, a robust and heterogeneous evaluation benchmark for information retrieval. We evaluate on all tasks in this benchmark that are publicly available~\citep{kamalloo2023resources}. For all datasets we use the the test set, except for MSMARCO where we follow BEIR convention to evaluate on the dev set. For \textit{cqadupstack} we follow BEIR convention and evaluate on all sub-datasets and compute their average.
Because BEIR is a large benchmark, and the largest dataset has more than $13$K queries, we only generate query analysis to evaluate on this dataset. This almost adds no overhead to the generation because the query analysis generation does not depend on the document. Since our model only generates a single token "Yes" or "No", its latency is almost the same as encoding both the query and the document with an encoding-based retrieval model, making it a highly efficient alternative.

\subsection{First-stage retrieval}
For both benchmarks we evaluate on, we follow common settings from previous work to rerank the top-$100$ documents from the first-stage retrieval and use nDCG@10 score as the evaluation metric. This metric assesses the quality of the retrieved documents, taking into account both their relevance and ranking position.

\paragraph{BRIGHT}
The original BRIGHT paper explores using LLMs to generate Chain-of-Thought~\citep{wei2022chain} reasoning steps as queries~\citep{su2024bright}, resulting in up to $12.2$ point improvements on average. We thus build on top of this best first-retrieval model on the leaderboard,
which achieves an nDCG@10 score of $26.5$ with BM25 and reasoning chains generated by GPT-$4$-$0125$-preview.

\paragraph{BEIR}
We follow the original BEIR paper~\citep{thakur2021beir} and use ElasticSearch BM25~\citep{elasticsearch2018elasticsearch} as the first-stage retriever.

\subsection{Base model}
Our main model builds on top of Llama-3.1-70B-instruct~\citep{dubey2024llama}. We choose the Llama-3.1 model family because its \textsc{RoPE} scaling~\cite{liu2024scaling} allows longer context length up to $128$K, which is essential in handling long documents. To speed up experiments, we evaluate on a quantized version of this model, namely Llama-3.1-70B-instruct-awq-int4.\footnote{\url{https://huggingface.co/hugging-quants/Meta-Llama-3.1-70B-Instruct-AWQ-INT4}} We also perform ablation studies where we evaluate on Llama-3.1-8B\footnote{\url{https://huggingface.co/meta-llama/Llama-3.1-8B}} and Llama-3.1-405B-instruct-awq-int4.\footnote{\url{https://huggingface.co/hugging-quants/Meta-Llama-3.1-405B-Instruct-AWQ-INT4}} Note that for the 8B model we do not use the quantized version because it can already fit on a single A$100$ GPU, while the other two bigger models require quantization to save computational cost. More specifically, the $70$B version requires at least $2$ x A$100$ GPUs, while the $405$B version requires $8$.

\subsection{Baseline rerankers}
We reproduce \textsc{RankLlama}~\citep{10.1145/3626772.3657951} and \textsc{RankZephyr}~\citep{pradeep2023rankzephyr}, two state-of-the-art rerankers as evaluated by the BEIR benchmark. RankLlama is a pointwise reranker that directly outputs a score. This model is trained on the MS MARCO passage ranking dataset~\citep{bajaj2016ms}. \textsc{RankZephyr} is a listwise reranker that takes a query and a list of documents together as input and outputs a ranking. This model uses the queries sourced by~\citet{sun-etal-2023-chatgpt} from the MS MARCO dataset to distill GPT-3.5 and GPT-4 in sequence. We use the \textsc{rerankers} library~\citep{clavié2024rerankers}, a lightweight unified API that allows users to run diverse reranking models out-of-the-box.\footnote{\url{https://github.com/AnswerDotAI/rerankers}}

\subsection{Efficiency and Optimization}
To make the encoding and generation more efficient, we use vLLM~\citep{kwon2023efficient}. This library leverages paged attention, which improves the throughput of popular LLMs by $2$ - $4$x with the same level of latency. Importantly, when designing the prompts used in our approach, we append the query and the document at the very end of each prompt to make the best use of Automatic Prefix Caching~\citep{gim2024prompt}, which temporarily stores the KV cache of existing inputs so that a new input can directly reuse the KV cache if it shares the same prefix with one of the existing ones. This allows the new inputs to skip the computation of the shared part. This design greatly improves the efficiency of our experiments.
\section{Results and Discussion}
\subsection{Main Results}
\paragraph{BRIGHT}
As shown in Table~\ref{tab:bright}, \textsc{JudgeRank} achieves state-of-the-art results on the BRIGHT evaluation benchmark as measured by nDCG@10. Our best preforming model improves upon the no-rerank baseline by more than $9$ points, while \textsc{RankLlama} underperforms the baseline and \textsc{RankZephyr} stays barely above the baseline. The smaller Llama-$3.1$-$8$B-instruct also outperforms the baseline by more than $3$ points, showing the generalizability of our approach across different model sizes. Interestingly, increasing model size from $70$B to $405$B does not bring a significant gain on nDCG@10. We thus select the $70$B version as our main model to balance between efficiency and performance.
In contrast, the original BRIGHT paper shows that GPT-4 with listwise reranking improves on top of vanilla BM25 baseline by an average of $2.7$ points on nDCG@10, a much smaller improvement than our approach despite using a much stronger LLM.


\setlength{\tabcolsep}{4.3pt}
\begin{table*}[t]
\centering
\footnotesize
\medskip
\begin{tabular}{rcccccccc}
\cmidrule[0.8pt]{2-8}
&\multirow{2}{*}{\;\; BM25 \;\;} &\multirow{2}{*}{RankLlama} &\multirow{2}{*}{RankZephyr} &\multicolumn{4}{c}{JudgeRank} \\\cmidrule{5-8}
& & & &\;\; 8B \;\; &\;\; 70B \;\;&\; 405B \; &Ensemble \\
\midrule
Biology &53.63 &11.05 &44.37 &54.44 &57.59 &60.33 &\textbf{60.70} \\
Earth Science &53.65 &11.83 &35.24 &54.62 &58.36 &55.11 &\textbf{58.72} \\
Economics &24.28 &\: 9.56 &24.44 &28.04 &33.01 &32.15 & \textbf{35.39} \\
Psychology &38.59 &11.38 &36.92 &42.16 &46.59 &45.42 & \textbf{47.57} \\
Robotics &18.77 &\: 8.53 &18.80 &24.09 &\textbf{28.30} &27.64 &28.16 \\
Stack Overflow &22.74 &11.40 &19.62 &27.18 &27.47 &28.30 &\textbf{29.74} \\
Sustainable Living &25.90 &11.75 &29.38 &30.69 &39.55 &38.54 &\textbf{41.88} \\
Leetcode &19.27 &20.32 &\textbf{24.58} &17.49 &20.06 &22.47 &20.23 \\
Pony &17.73 &18.88 &\textbf{48.96} &22.85 &30.82 &31.54 &32.74 \\
Aops &\: 3.92 &\: 3.55 &\: 6.98 &\: 6.15 &\: 8.24 &\: 7.74 &\: \textbf{8.57} \\
TheoremQA-Questions &18.90 &11.82 &22.34 &24.09 &23.83 &\textbf{26.93} &25.86 \\
TheoremQA-Theorems &20.22 &\: 5.63 &\: 7.78 &29.89 &34.55 &36.16 &\textbf{36.20} \\
\midrule
Average &26.47 &11.31 &27.25 &30.14 &34.03 &34.36 &\textbf{35.48} \\
\bottomrule
\end{tabular}
\caption{\textsc{JudgeRank} nDCG@10 results on the BRIGHT evaluation benchmark. Best results on each dataset and the entire benchmark are boldfaced. "Ensemble" stands for model ensembling of \textsc{JudgeRank}-$8$B, $70$B, and $405$B.}
\label{tab:bright}
\end{table*}

\paragraph{BEIR}
As shown in Table~\ref{tab:beir}, our model delivers competitive results on the BEIR evaluation benchmark despite the fact that \textsc{RankLlama} and \textsc{RankZephyr} are heavily fine-tuned on in-domain data including MS MARCO, which is part of the BEIR benchmark.

\begin{table*}[t]
\centering
\footnotesize
\medskip
\begin{tabular}{rccccc}
\cmidrule[0.8pt]{2-5}
&\;\; BM25 \;\; &JudgeRank &RankLlama &RankZephyr \\
\midrule
webis-touche2020 &\textbf{34.71} &27.94 &32.97 &33.34 \\
trec-covid &68.80 &83.70 &81.08 &\textbf{85.77} \\
scifact &69.06 &73.24 &75.57 &\textbf{74.68} \\
nfcorpus &34.28 &37.73 &35.92 &\textbf{38.79} \\
dbpedia-entity &32.02 &\textbf{44.30} &43.72 &44.19 \\
fiqa &25.36 &40.35 &\textbf{42.70} &40.40 \\
scidocs &16.47 &19.47 &19.26 &\textbf{19.70} \\
arguana &47.16 &\textbf{62.77} &32.08 &47.07 \\
nq &32.61 &56.87 &59.19 &\textbf{60.10} \\
climate-fever &18.61 &19.00 &17.75 &\textbf{24.26} \\
fever &64.94 &69.97 &78.55 &\textbf{79.52} \\
msmarco &22.75 &34.01 &\textbf{41.40} &37.89 \\
hotpotqa &60.22 &68.30 &70.65 &\textbf{71.16} \\
quora &80.77 &\textbf{84.25} &82.73 &79.60 \\
cqadupstack &32.53 &\textbf{43.98} & 42.43 &42.68 \\
\midrule
Average &42.69 &51.06 &50.40 &\textbf{51.94} \\
\bottomrule
\end{tabular}
\caption{\textsc{JudgeRank} nDCG@10 results on the BEIR evaluation benchmark. Best results on each dataset and the entire benchmark are boldfaced.}
\label{tab:beir}
\end{table*}

\subsection{Discussion}
\label{subsec:analysis}
We pose several research questions to illustrate whether and how our approach works.

\begin{figure*}[t]
\centering
    \begin{subfigure}[t]{0.49\textwidth}
    \centering
    \includegraphics[width=1.0\textwidth]{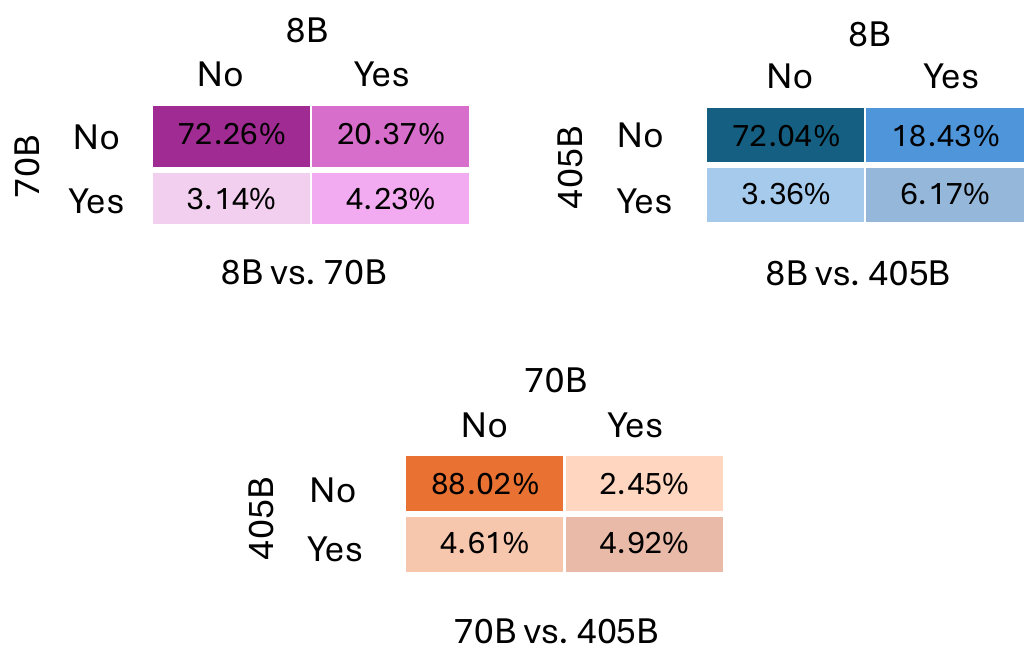}
    \end{subfigure}
    \hfill
    \begin{subfigure}[t]{0.49\textwidth}
    \centering
    \includegraphics[width=1.0\textwidth]{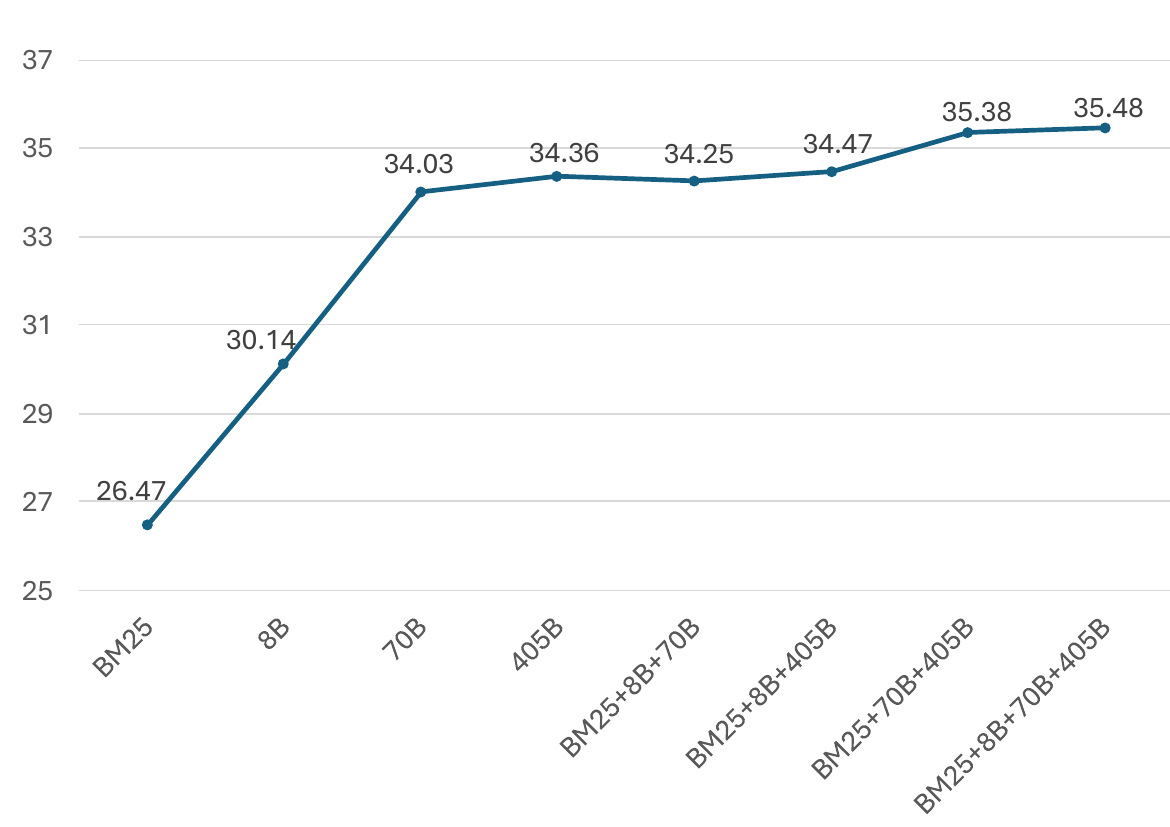}
    \end{subfigure}
    \caption{\textbf{On the left}: judgment alignment studies for models of three sizes: $8$B, $70$B, and $405$B. Percentages are shown for each quadrant. \textbf{On the right}: nDCG@10 of each individual model and model ensembling on the BRIGHT evaluation benchmark.}
    \label{fig:ensemble}
\end{figure*}

\paragraph{How complementary are LLMs of different scales?}
\label{subsec:comparison-70b-405b}
In Table~\ref{tab:bright}, we observe that JudgeRank-$70$B and JudgeRank-$405$B performs on par with each other. However, nDCG@10 alone does not reveal the whole picture. One natural question to ask is: do these two models make similar judgments or are complementary to each other? To answer this question, we obtain statistics on the percentage of both models agreeing and disagreeing each other and show them on the left of Figure~\ref{fig:ensemble}. From the tables we can see that for all three combinations of the models, the majority case is always that both models rejects the documents. This is understandable because only a few out of the top-$100$ documents are supposed to be relevant. The interesting pattern emerges when we inspect the other three cases: each pair of the models spends more time disagreeing with each other than both outputting "Yes". For the pairs $8$B vs $70$B and $8$B vs $405$B, there is a higher difference because the capabilities of the two models differ more. In contrast, for $70$B vs $405$B there is less disagreement. From these observations, we indeed see that each two models may be complementary to each other.

Motivated by this observation, we take model ensembling one step further. Recall that in Section~\ref{subsec:scoring}, we ensemble the BM25 score with each of the scores output by the Llama models. Here we first take the average score output by all the Llama models, and then perform the weighted sum with the BM25 score. More specifically, let $S_{8\text{B}}$, $S_{70\text{B}}$, and $S_{405\text{B}}$ be the score assigned by each model, respectively, the ensemble score of the three models is computed as $\alpha (S_{8\text{B}} + S_{70\text{B}} + S_{405\text{B}}) / 3 + S_{\text{BM25}}$, where again $\alpha = 100$ and $S_{\text{BM25}}$ is the score given by the BM25 model. The same equation generalizes analogously to two-model ensembles.

We present all ensembling results on the right of Figure~\ref{fig:ensemble}. We can see that each ensembling performance is better than its individual model performances, with the strongest performance observed when ensembling all three models. This result shows that a salient performance boost can be achieved by ensembling two of the strongest models ($70$B + $405$B), while even the model with lower performance (i.e., $8$B) could contribute positively in model ensembling. Intuitively, such ensembling is equivalent to a verification or a majority voting step. The final score is the highest when both models say "Yes", the score is medium when one of the two says "No", and the lowest score is observed when both say "No".

\begin{figure*}[t]
\centering
    \begin{subfigure}[t]{0.49\textwidth}
    \centering
    \includegraphics[width=1.0\textwidth]{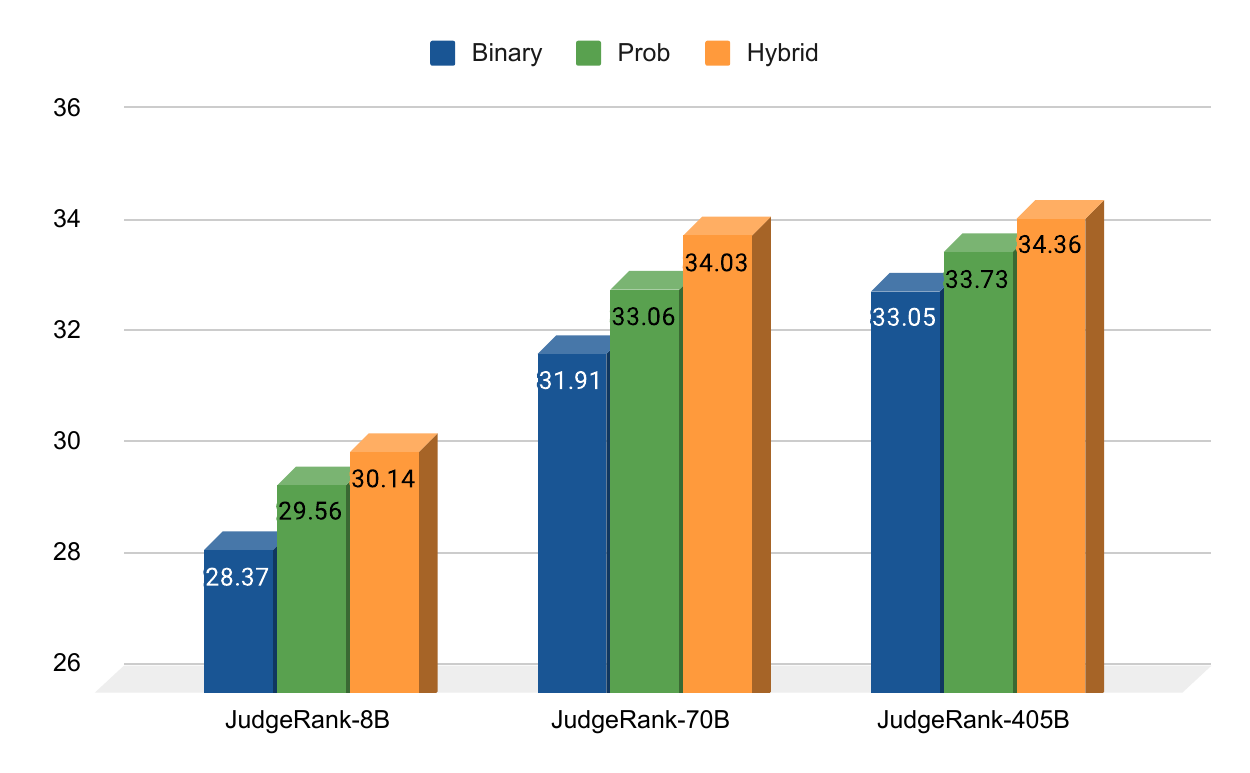}
    \end{subfigure}
    \hfill
    \begin{subfigure}[t]{0.49\textwidth}
    \centering
    \includegraphics[width=1.0\textwidth]{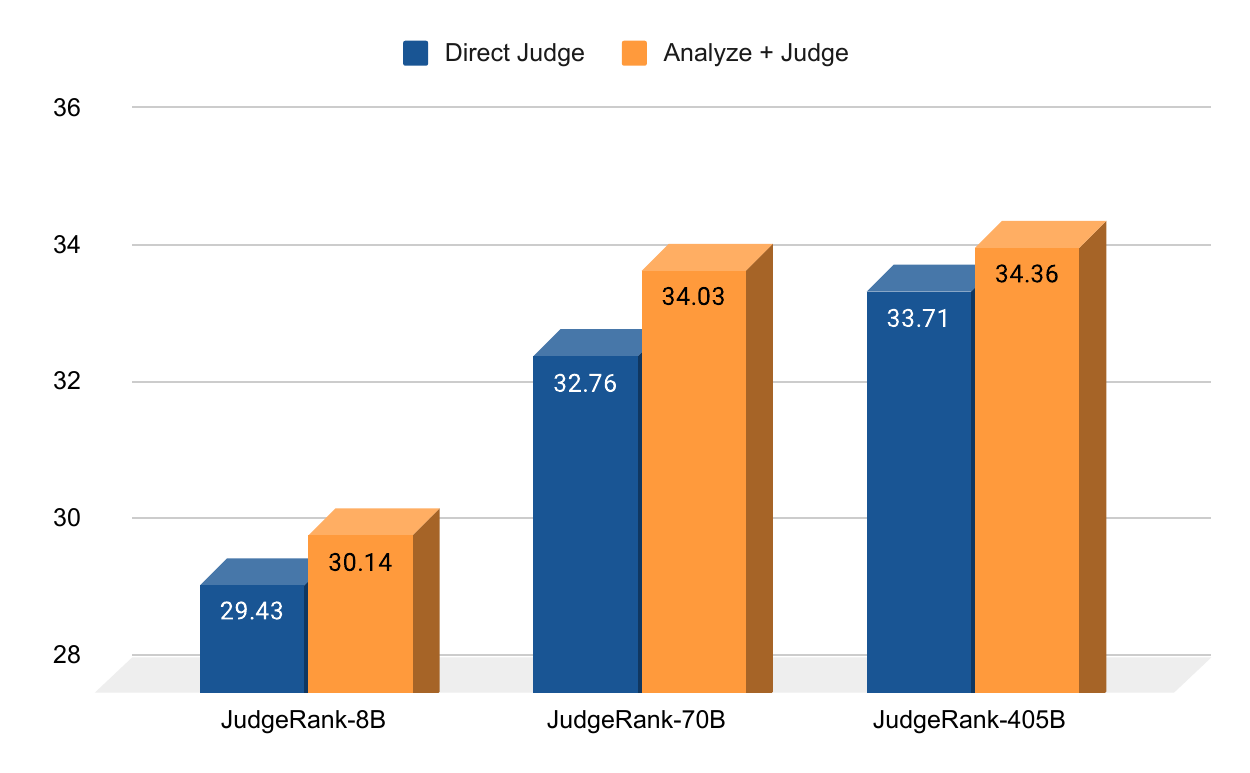}
    \end{subfigure}
    \caption{Ablation studies of JudgeRank. \textbf{On the left}: Comparison of three scoring settings on the BRIGHT evaluation benchmark. \textit{Binary} stands for binary judgment, \textit{Prob} stands for probability, and \textit{Hybrid} stands for a weighted sum of BM25 and probability scores. \textbf{On the right}: Comparison of direct judge and judge with query and document analyses on the BRIGHT evaluation benchmark.}
    \label{fig:ablation}
\end{figure*}

\paragraph{How does the choice of reranking score impact the final performance?}
Recall that to compute the final score, we take a weighted sum of the BM25 score and the probability score from the judgment step. To justify this choice, we compare it with two other settings: the first is binary judgment, and the second only uses the normalized probability to rerank documents (introduced in Section~\ref{sec:experimental-setup}). The left part of Figure~\ref{fig:ablation} shows that binary judgment performs the worst among the three settings while using only probability achieves somewhere in between. 
This is understandable because binary judgments are sensitive to wordings. Imagine that if we change the relation from "substantially helps answer" to "helps answer" or "at least partially helps answer," the number of "Yes" that the model outputs will keep increasing, thus also increasing the number of false positives. However, the other two settings are not sensitive to such changes.

\paragraph{How useful are the query and document analysis steps?}
To show the effectiveness of the analysis steps, we perform an ablation study on BRIGHT. We remove the two analyses steps and keep the judgment step untouched to compare its performance with the original approach. 
The right of Figure~\ref{fig:ablation} shows that judging with query and document analyses performs consistently better than the direct judgment approach.

\begin{figure*}[t]
\centering
\includegraphics[width=0.98\textwidth]{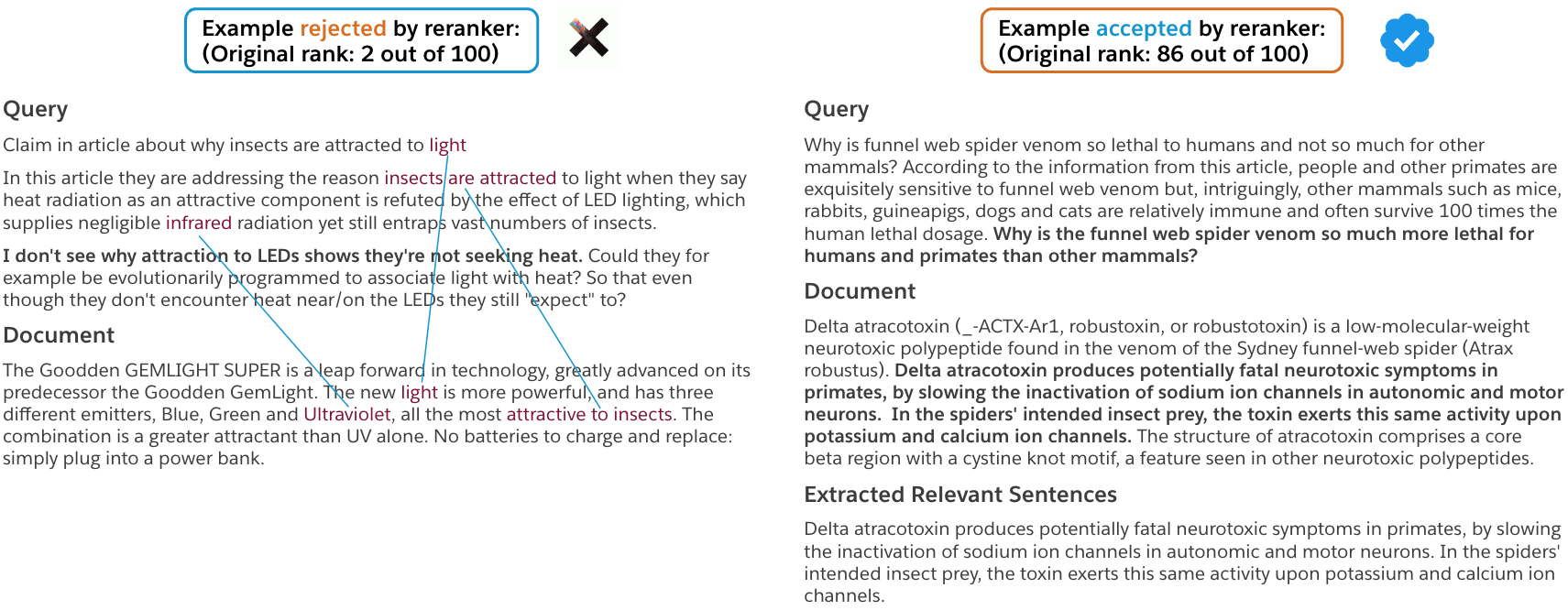}
\caption{Illustration of how agentic generations of \textsc{JudgeRank} help identifying the relevant documents. \textbf{On the left}, the document is ranked high by the first-stage retriever but judged as negative by the reranker. \textbf{On the right}, the document is ranked low by the first-stage retriever but judged as positive by the reranker because the document analysis prompt helps the LLM to locate the relevant sentences that answer the query.}
\label{fig:extractive-summary}
\end{figure*}

\paragraph{Qualitative examples}
Figure~\ref{fig:extractive-summary} demonstrates how \textsc{JudgeRank} enhances document relevance identification using real examples from the BRIGHT dataset. In the left example, we observe a document initially ranked high by the first-stage retriever due to significant word overlap with the query. However, \textsc{JudgeRank}'s deeper analysis correctly judges this document as irrelevant. Despite surface-level similarities, the reranker fails to extract any sentences that help answer the query, revealing that the document is merely an advertisement coincidentally sharing common terminology with the query. The right example presents a contrasting scenario, where a document initially ranked low by the first-stage retriever due to minimal word overlap with the query is accurately identified by \textsc{JudgeRank} as highly relevant. In this instance, the document analysis prompt enables the LLM to pinpoint key sentences that elucidate the underlying mechanism of funnel web spider venom's lethality, precisely addressing the query's intent. These extracted sentences further inform the LLM to make the final positive judgment, demonstrating JudgeRank's ability to uncover deeply relevant content that might be overlooked by traditional retrieval methods.

\section{Related Work}
The field of reranking models can be understood through two primary dimensions. 

The first dimension distinguishes between encoding-based and LLM-based approaches. Encoding-based models~\citep{nogueira2019passage,gao2021rethink}, which have been the standard since the introduction of BERT, typically require extensive training to adapt to the specific objectives of reranking tasks. In contrast, LLM-based models~\citep{sun-etal-2023-chatgpt,tang-etal-2024-found,qin-etal-2024-large,zhuang2024setwise}, particularly those with decoder-only architectures, often demonstrate impressive zero-shot capabilities, allowing them to perform effectively without task-specific fine-tuning. However, it's worth noting that some researchers have explored pretraining, fine-tuning and distillation techniques for LLM-based rerankers to further enhance their performance~\citep{zhang2023rankinggpt,10.1145/3626772.3657951,yu2024rankrag}.

The second dimension differentiates between pointwise and listwise reranking strategies. Pointwise rerankers~\citep{10.1145/3626772.3657951,guo2024generating} evaluate the relevance of individual query-document pairs in isolation, producing a score for each pair without direct comparison between documents. Listwise rerankers~\citep{sun-etal-2023-chatgpt,pradeep2023rankzephyr,ma2023zero,yoon-etal-2024-listt5,liu2024leveraging,tang-etal-2024-found}, on the other hand, consider the entire set of documents for a given query, generating a fully ordered list as output. To address the challenges posed by input length limitations, researchers have developed innovative techniques such as sliding window approaches~\citep{sun-etal-2023-chatgpt}, which allow for the ranking of smaller subsets of documents before aggregating them into a comprehensive ranking. This framework encompasses pairwise~\citep{pradeep2021expando,qin-etal-2024-large} and setwise~\citep{zhuang2024setwise} rerankers as specific instances of the broader listwise category.

Our contribution is a novel LLM-based, pointwise reranker that leverages the capabilities of instruction-tuned language models and incorporates explicit reasoning steps in the relevance judgment process. This approach sets our work apart from previous efforts by enhancing the model's ability to handle complex, reasoning-intensive reranking tasks while simultaneously improving the interpretability of its decisions.

\section{Conclusion}
\label{sec:conclusion}
In this work, we target document retrieval tasks that require intensive context-based reasoning, which even the strongest retrieval models struggle to achieve satisfactory performance. Through experiments and ablation studies, we show that our agentic reranker can effectively recover low-ranked documents and outperform previous state-of-the-art reranking models, while remaining flexible and efficient. In section~\ref{subsec:comparison-70b-405b}, we have shown the significant benefit of model ensembling on document reranking; yet we do not have to stop there. There are at least two categories of ensembling that we envision. 
First, sampling ensembling: For each generation, we sample several generations, each of which could lead to a different judgment. This kind of ensembling is similar to the self-consistency approach~\citep{tang-etal-2024-found}.
Second, prompt ensembling: we could leverage paraphrases of the same prompt to perform ensembling.
We leave the exploration as future work because such approaches could be generalized to many prompting-related tasks, and thus better be addressed in separate works.



\bibliography{custom}

\begin{thebibliography}{40}
\providecommand{\natexlab}[1]{#1}

\bibitem[{Bajaj et~al.(2016)Bajaj, Campos, Craswell, Deng, Gao, Liu, Majumder, McNamara, Mitra, Nguyen et~al.}]{bajaj2016ms}
Payal Bajaj, Daniel Campos, Nick Craswell, Li~Deng, Jianfeng Gao, Xiaodong Liu, Rangan Majumder, Andrew McNamara, Bhaskar Mitra, Tri Nguyen, et~al. 2016.
\newblock Ms marco: A human generated machine reading comprehension dataset.
\newblock \emph{arXiv preprint arXiv:1611.09268}.

\bibitem[{Clavié(2024)}]{clavié2024rerankers}
Benjamin Clavié. 2024.
\newblock \href {https://arxiv.org/abs/2408.17344} {rerankers: A lightweight python library to unify ranking methods}.
\newblock \emph{Preprint}, arXiv:2408.17344.

\bibitem[{Dubey et~al.(2024)Dubey, Jauhri, Pandey, Kadian, Al-Dahle, Letman, Mathur, Schelten, Yang, Fan et~al.}]{dubey2024llama}
Abhimanyu Dubey, Abhinav Jauhri, Abhinav Pandey, Abhishek Kadian, Ahmad Al-Dahle, Aiesha Letman, Akhil Mathur, Alan Schelten, Amy Yang, Angela Fan, et~al. 2024.
\newblock The llama 3 herd of models.
\newblock \emph{arXiv preprint arXiv:2407.21783}.

\bibitem[{Elasticsearch(2018)}]{elasticsearch2018elasticsearch}
BV~Elasticsearch. 2018.
\newblock Elasticsearch.
\newblock \emph{software], version}, 6(1).

\bibitem[{Gao et~al.(2021)Gao, Dai, and Callan}]{gao2021rethink}
Luyu Gao, Zhuyun Dai, and Jamie Callan. 2021.
\newblock Rethink training of bert rerankers in multi-stage retrieval pipeline.
\newblock In \emph{Advances in Information Retrieval: 43rd European Conference on IR Research, ECIR 2021, Virtual Event, March 28--April 1, 2021, Proceedings, Part II 43}, pages 280--286. Springer.

\bibitem[{Gim et~al.(2024)Gim, Chen, Lee, Sarda, Khandelwal, and Zhong}]{gim2024prompt}
In~Gim, Guojun Chen, Seung-seob Lee, Nikhil Sarda, Anurag Khandelwal, and Lin Zhong. 2024.
\newblock Prompt cache: Modular attention reuse for low-latency inference.
\newblock \emph{Proceedings of Machine Learning and Systems}, 6:325--338.

\bibitem[{Guo et~al.(2024)Guo, Li, Zhuang, Luo, Li, Yan, and Zhang}]{guo2024generating}
Fang Guo, Wenyu Li, Honglei Zhuang, Yun Luo, Yafu Li, Le~Yan, and Yue Zhang. 2024.
\newblock Generating diverse criteria on-the-fly to improve point-wise llm rankers.
\newblock \emph{arXiv preprint arXiv:2404.11960}.

\bibitem[{Izacard et~al.(2022)Izacard, Caron, Hosseini, Riedel, Bojanowski, Joulin, and Grave}]{izacard2022unsupervised}
Gautier Izacard, Mathilde Caron, Lucas Hosseini, Sebastian Riedel, Piotr Bojanowski, Armand Joulin, and Edouard Grave. 2022.
\newblock \href {https://openreview.net/forum?id=jKN1pXi7b0} {Unsupervised dense information retrieval with contrastive learning}.
\newblock \emph{Transactions on Machine Learning Research}.

\bibitem[{Kamalloo et~al.(2023)Kamalloo, Thakur, Lassance, Ma, Yang, and Lin}]{kamalloo2023resources}
Ehsan Kamalloo, Nandan Thakur, Carlos Lassance, Xueguang Ma, Jheng-Hong Yang, and Jimmy Lin. 2023.
\newblock \href {https://arxiv.org/abs/2306.07471} {Resources for brewing beir: Reproducible reference models and an official leaderboard}.
\newblock \emph{Preprint}, arXiv:2306.07471.

\bibitem[{Kwon et~al.(2023)Kwon, Li, Zhuang, Sheng, Zheng, Yu, Gonzalez, Zhang, and Stoica}]{kwon2023efficient}
Woosuk Kwon, Zhuohan Li, Siyuan Zhuang, Ying Sheng, Lianmin Zheng, Cody~Hao Yu, Joseph Gonzalez, Hao Zhang, and Ion Stoica. 2023.
\newblock Efficient memory management for large language model serving with pagedattention.
\newblock In \emph{Proceedings of the 29th Symposium on Operating Systems Principles}, pages 611--626.

\bibitem[{Lewis et~al.(2020)Lewis, Perez, Piktus, Petroni, Karpukhin, Goyal, K\"{u}ttler, Lewis, Yih, Rockt\"{a}schel, Riedel, and Kiela}]{LewisRAG}
Patrick Lewis, Ethan Perez, Aleksandra Piktus, Fabio Petroni, Vladimir Karpukhin, Naman Goyal, Heinrich K\"{u}ttler, Mike Lewis, Wen-tau Yih, Tim Rockt\"{a}schel, Sebastian Riedel, and Douwe Kiela. 2020.
\newblock Retrieval-augmented generation for knowledge-intensive nlp tasks.
\newblock In \emph{Proceedings of the 34th International Conference on Neural Information Processing Systems}, NIPS '20, Red Hook, NY, USA. Curran Associates Inc.

\bibitem[{Liang et~al.(2023)Liang, Bommasani, Lee, Tsipras, Soylu, Yasunaga, Zhang, Narayanan, Wu, Kumar, Newman, Yuan, Yan, Zhang, Cosgrove, Manning, Re, Acosta-Navas, Hudson, Zelikman, Durmus, Ladhak, Rong, Ren, Yao, WANG, Santhanam, Orr, Zheng, Yuksekgonul, Suzgun, Kim, Guha, Chatterji, Khattab, Henderson, Huang, Chi, Xie, Santurkar, Ganguli, Hashimoto, Icard, Zhang, Chaudhary, Wang, Li, Mai, Zhang, and Koreeda}]{liang2023holistic}
Percy Liang, Rishi Bommasani, Tony Lee, Dimitris Tsipras, Dilara Soylu, Michihiro Yasunaga, Yian Zhang, Deepak Narayanan, Yuhuai Wu, Ananya Kumar, Benjamin Newman, Binhang Yuan, Bobby Yan, Ce~Zhang, Christian~Alexander Cosgrove, Christopher~D Manning, Christopher Re, Diana Acosta-Navas, Drew~Arad Hudson, Eric Zelikman, Esin Durmus, Faisal Ladhak, Frieda Rong, Hongyu Ren, Huaxiu Yao, Jue WANG, Keshav Santhanam, Laurel Orr, Lucia Zheng, Mert Yuksekgonul, Mirac Suzgun, Nathan Kim, Neel Guha, Niladri~S. Chatterji, Omar Khattab, Peter Henderson, Qian Huang, Ryan~Andrew Chi, Sang~Michael Xie, Shibani Santurkar, Surya Ganguli, Tatsunori Hashimoto, Thomas Icard, Tianyi Zhang, Vishrav Chaudhary, William Wang, Xuechen Li, Yifan Mai, Yuhui Zhang, and Yuta Koreeda. 2023.
\newblock \href {https://openreview.net/forum?id=iO4LZibEqW} {Holistic evaluation of language models}.
\newblock \emph{Transactions on Machine Learning Research}.
\newblock Featured Certification, Expert Certification.

\bibitem[{Liu et~al.(2024{\natexlab{a}})Liu, Wang, Wang, and Mao}]{liu2024leveraging}
Qi~Liu, Bo~Wang, Nan Wang, and Jiaxin Mao. 2024{\natexlab{a}}.
\newblock Leveraging passage embeddings for efficient listwise reranking with large language models.
\newblock \emph{arXiv preprint arXiv:2406.14848}.

\bibitem[{Liu et~al.(2024{\natexlab{b}})Liu, Yan, An, Qiu, and Lin}]{liu2024scaling}
Xiaoran Liu, Hang Yan, Chenxin An, Xipeng Qiu, and Dahua Lin. 2024{\natexlab{b}}.
\newblock \href {https://openreview.net/forum?id=JO7k0SJ5V6} {Scaling laws of ro{PE}-based extrapolation}.
\newblock In \emph{The Twelfth International Conference on Learning Representations}.

\bibitem[{Ma et~al.(2024)Ma, Wang, Yang, Wei, and Lin}]{10.1145/3626772.3657951}
Xueguang Ma, Liang Wang, Nan Yang, Furu Wei, and Jimmy Lin. 2024.
\newblock \href {https://doi.org/10.1145/3626772.3657951} {Fine-tuning llama for multi-stage text retrieval}.
\newblock In \emph{Proceedings of the 47th International ACM SIGIR Conference on Research and Development in Information Retrieval}, SIGIR '24, page 2421–2425, New York, NY, USA. Association for Computing Machinery.

\bibitem[{Ma et~al.(2023)Ma, Zhang, Pradeep, and Lin}]{ma2023zero}
Xueguang Ma, Xinyu Zhang, Ronak Pradeep, and Jimmy Lin. 2023.
\newblock Zero-shot listwise document reranking with a large language model.
\newblock \emph{arXiv preprint arXiv:2305.02156}.

\bibitem[{Masson(1983)}]{masson1983conceptual}
Michael~EJ Masson. 1983.
\newblock Conceptual processing of text during skimming and rapid sequential reading.
\newblock \emph{Memory \& cognition}, 11(3):262--274.

\bibitem[{Muennighoff et~al.(2022)Muennighoff, Tazi, Magne, and Reimers}]{muennighoff2022mteb}
Niklas Muennighoff, Nouamane Tazi, Lo{\"\i}c Magne, and Nils Reimers. 2022.
\newblock \href {https://doi.org/10.48550/ARXIV.2210.07316} {Mteb: Massive text embedding benchmark}.
\newblock \emph{arXiv preprint arXiv:2210.07316}.

\bibitem[{Nguyen(2016)}]{nguyen2016ms}
T~Nguyen. 2016.
\newblock Ms marco: A human generated machine reading comprehension dataset.
\newblock \emph{arXiv preprint arXiv:1611.09268}.

\bibitem[{Nogueira and Cho(2019)}]{nogueira2019passage}
Rodrigo Nogueira and Kyunghyun Cho. 2019.
\newblock Passage re-ranking with bert.
\newblock \emph{arXiv preprint arXiv:1901.04085}.

\bibitem[{Nogueira et~al.(2020)Nogueira, Jiang, Pradeep, and Lin}]{nogueira-etal-2020-document}
Rodrigo Nogueira, Zhiying Jiang, Ronak Pradeep, and Jimmy Lin. 2020.
\newblock \href {https://doi.org/10.18653/v1/2020.findings-emnlp.63} {Document ranking with a pretrained sequence-to-sequence model}.
\newblock In \emph{Findings of the Association for Computational Linguistics: EMNLP 2020}, pages 708--718, Online. Association for Computational Linguistics.

\bibitem[{Ouyang et~al.(2022)Ouyang, Wu, Jiang, Almeida, Wainwright, Mishkin, Zhang, Agarwal, Slama, Ray et~al.}]{ouyang2022training}
Long Ouyang, Jeffrey Wu, Xu~Jiang, Diogo Almeida, Carroll Wainwright, Pamela Mishkin, Chong Zhang, Sandhini Agarwal, Katarina Slama, Alex Ray, et~al. 2022.
\newblock Training language models to follow instructions with human feedback.
\newblock \emph{Advances in neural information processing systems}, 35:27730--27744.

\bibitem[{Pradeep et~al.(2021)Pradeep, Nogueira, and Lin}]{pradeep2021expando}
Ronak Pradeep, Rodrigo Nogueira, and Jimmy Lin. 2021.
\newblock The expando-mono-duo design pattern for text ranking with pretrained sequence-to-sequence models.
\newblock \emph{arXiv preprint arXiv:2101.05667}.

\bibitem[{Pradeep et~al.(2023)Pradeep, Sharifymoghaddam, and Lin}]{pradeep2023rankzephyr}
Ronak Pradeep, Sahel Sharifymoghaddam, and Jimmy Lin. 2023.
\newblock {RankZephyr}: Effective and robust zero-shot listwise reranking is a breeze!
\newblock \emph{arXiv:2312.02724}.

\bibitem[{Qin et~al.(2024)Qin, Jagerman, Hui, Zhuang, Wu, Yan, Shen, Liu, Liu, Metzler, Wang, and Bendersky}]{qin-etal-2024-large}
Zhen Qin, Rolf Jagerman, Kai Hui, Honglei Zhuang, Junru Wu, Le~Yan, Jiaming Shen, Tianqi Liu, Jialu Liu, Donald Metzler, Xuanhui Wang, and Michael Bendersky. 2024.
\newblock \href {https://doi.org/10.18653/v1/2024.findings-naacl.97} {Large language models are effective text rankers with pairwise ranking prompting}.
\newblock In \emph{Findings of the Association for Computational Linguistics: NAACL 2024}, pages 1504--1518, Mexico City, Mexico. Association for Computational Linguistics.

\bibitem[{Robertson et~al.(2009)Robertson, Zaragoza et~al.}]{robertson2009probabilistic}
Stephen Robertson, Hugo Zaragoza, et~al. 2009.
\newblock The probabilistic relevance framework: Bm25 and beyond.
\newblock \emph{Foundations and Trends{\textregistered} in Information Retrieval}, 3(4):333--389.

\bibitem[{Robertson et~al.(1995)Robertson, Walker, Jones, Hancock-Beaulieu, Gatford et~al.}]{robertson1995okapi}
Stephen~E Robertson, Steve Walker, Susan Jones, Micheline~M Hancock-Beaulieu, Mike Gatford, et~al. 1995.
\newblock Okapi at trec-3.
\newblock \emph{Nist Special Publication Sp}, 109:109.

\bibitem[{Su et~al.(2024)Su, Yen, Xia, Shi, Muennighoff, Wang, Liu, Shi, Siegel, Tang et~al.}]{su2024bright}
Hongjin Su, Howard Yen, Mengzhou Xia, Weijia Shi, Niklas Muennighoff, Han-yu Wang, Haisu Liu, Quan Shi, Zachary~S Siegel, Michael Tang, et~al. 2024.
\newblock Bright: A realistic and challenging benchmark for reasoning-intensive retrieval.
\newblock \emph{arXiv preprint arXiv:2407.12883}.

\bibitem[{Sun et~al.(2023)Sun, Yan, Ma, Wang, Ren, Chen, Yin, and Ren}]{sun-etal-2023-chatgpt}
Weiwei Sun, Lingyong Yan, Xinyu Ma, Shuaiqiang Wang, Pengjie Ren, Zhumin Chen, Dawei Yin, and Zhaochun Ren. 2023.
\newblock \href {https://doi.org/10.18653/v1/2023.emnlp-main.923} {Is {C}hat{GPT} good at search? investigating large language models as re-ranking agents}.
\newblock In \emph{Proceedings of the 2023 Conference on Empirical Methods in Natural Language Processing}, pages 14918--14937, Singapore. Association for Computational Linguistics.

\bibitem[{Tang et~al.(2024)Tang, Zhang, Ma, Lin, and Ture}]{tang-etal-2024-found}
Raphael Tang, Crystina Zhang, Xueguang Ma, Jimmy Lin, and Ferhan Ture. 2024.
\newblock \href {https://doi.org/10.18653/v1/2024.naacl-long.129} {Found in the middle: Permutation self-consistency improves listwise ranking in large language models}.
\newblock In \emph{Proceedings of the 2024 Conference of the North American Chapter of the Association for Computational Linguistics: Human Language Technologies (Volume 1: Long Papers)}, pages 2327--2340, Mexico City, Mexico. Association for Computational Linguistics.

\bibitem[{Team et~al.(2023)Team, Anil, Borgeaud, Wu, Alayrac, Yu, Soricut, Schalkwyk, Dai, Hauth et~al.}]{team2023gemini}
Gemini Team, Rohan Anil, Sebastian Borgeaud, Yonghui Wu, Jean-Baptiste Alayrac, Jiahui Yu, Radu Soricut, Johan Schalkwyk, Andrew~M Dai, Anja Hauth, et~al. 2023.
\newblock Gemini: a family of highly capable multimodal models.
\newblock \emph{arXiv preprint arXiv:2312.11805}.

\bibitem[{Thakur et~al.(2021)Thakur, Reimers, R\"{u}ckl\'{e}, Srivastava, and Gurevych}]{thakur2021beir}
Nandan Thakur, Nils Reimers, Andreas R\"{u}ckl\'{e}, Abhishek Srivastava, and Iryna Gurevych. 2021.
\newblock \href {https://datasets-benchmarks-proceedings.neurips.cc/paper_files/paper/2021/file/65b9eea6e1cc6bb9f0cd2a47751a186f-Paper-round2.pdf} {Beir: A heterogeneous benchmark for zero-shot evaluation of information retrieval models}.
\newblock In \emph{Proceedings of the Neural Information Processing Systems Track on Datasets and Benchmarks}, volume~1.

\bibitem[{Wang et~al.(2020)Wang, Wei, Dong, Bao, Yang, and Zhou}]{wang2020minilm}
Wenhui Wang, Furu Wei, Li~Dong, Hangbo Bao, Nan Yang, and Ming Zhou. 2020.
\newblock Minilm: Deep self-attention distillation for task-agnostic compression of pre-trained transformers.
\newblock \emph{Advances in Neural Information Processing Systems}, 33:5776--5788.

\bibitem[{Wei et~al.(2022)Wei, Wang, Schuurmans, Bosma, brian ichter, Xia, Chi, Le, and Zhou}]{wei2022chain}
Jason Wei, Xuezhi Wang, Dale Schuurmans, Maarten Bosma, brian ichter, Fei Xia, Ed~H. Chi, Quoc~V Le, and Denny Zhou. 2022.
\newblock \href {https://openreview.net/forum?id=_VjQlMeSB_J} {Chain of thought prompting elicits reasoning in large language models}.
\newblock In \emph{Advances in Neural Information Processing Systems}.

\bibitem[{Yoon et~al.(2024)Yoon, Choi, Kim, Yun, Kim, and Hwang}]{yoon-etal-2024-listt5}
Soyoung Yoon, Eunbi Choi, Jiyeon Kim, Hyeongu Yun, Yireun Kim, and Seung-won Hwang. 2024.
\newblock \href {https://aclanthology.org/2024.acl-long.125} {{L}ist{T}5: Listwise reranking with fusion-in-decoder improves zero-shot retrieval}.
\newblock In \emph{Proceedings of the 62nd Annual Meeting of the Association for Computational Linguistics (Volume 1: Long Papers)}, pages 2287--2308, Bangkok, Thailand. Association for Computational Linguistics.

\bibitem[{Yu et~al.(2024)Yu, Ping, Liu, Wang, You, Zhang, Shoeybi, and Catanzaro}]{yu2024rankrag}
Yue Yu, Wei Ping, Zihan Liu, Boxin Wang, Jiaxuan You, Chao Zhang, Mohammad Shoeybi, and Bryan Catanzaro. 2024.
\newblock Rankrag: Unifying context ranking with retrieval-augmented generation in llms.
\newblock \emph{arXiv preprint arXiv:2407.02485}.

\bibitem[{Zhang et~al.(2023)Zhang, Zhang, Long, Xie, Zhang, and Zhang}]{zhang2023rankinggpt}
Longhui Zhang, Yanzhao Zhang, Dingkun Long, Pengjun Xie, Meishan Zhang, and Min Zhang. 2023.
\newblock Rankinggpt: Empowering large language models in text ranking with progressive enhancement.
\newblock \emph{arXiv preprint arXiv:2311.16720}.

\bibitem[{Zheng et~al.(2023)Zheng, Chiang, Sheng, Zhuang, Wu, Zhuang, Lin, Li, Li, Xing, Zhang, Gonzalez, and Stoica}]{zheng2023judging}
Lianmin Zheng, Wei-Lin Chiang, Ying Sheng, Siyuan Zhuang, Zhanghao Wu, Yonghao Zhuang, Zi~Lin, Zhuohan Li, Dacheng Li, Eric Xing, Hao Zhang, Joseph~E. Gonzalez, and Ion Stoica. 2023.
\newblock \href {https://openreview.net/forum?id=uccHPGDlao} {Judging {LLM}-as-a-judge with {MT}-bench and chatbot arena}.
\newblock In \emph{Thirty-seventh Conference on Neural Information Processing Systems Datasets and Benchmarks Track}.

\bibitem[{Zhuang et~al.(2023)Zhuang, Qin, Jagerman, Hui, Ma, Lu, Ni, Wang, and Bendersky}]{zhuang2023rankt5}
Honglei Zhuang, Zhen Qin, Rolf Jagerman, Kai Hui, Ji~Ma, Jing Lu, Jianmo Ni, Xuanhui Wang, and Michael Bendersky. 2023.
\newblock Rankt5: Fine-tuning t5 for text ranking with ranking losses.
\newblock In \emph{Proceedings of the 46th International ACM SIGIR Conference on Research and Development in Information Retrieval}, pages 2308--2313.

\bibitem[{Zhuang et~al.(2024)Zhuang, Zhuang, Koopman, and Zuccon}]{zhuang2024setwise}
Shengyao Zhuang, Honglei Zhuang, Bevan Koopman, and Guido Zuccon. 2024.
\newblock A setwise approach for effective and highly efficient zero-shot ranking with large language models.
\newblock In \emph{Proceedings of the 47th International ACM SIGIR Conference on Research and Development in Information Retrieval}, pages 38--47.

\end{thebibliography}
\end{document}